\documentclass[runningheads]{llncs}

\usepackage{times}
\usepackage{soul}
\usepackage{url}
\usepackage[utf8]{inputenc}
\usepackage[small]{caption}
\usepackage{graphicx}
\usepackage{amsmath}
\usepackage{amsfonts}
\usepackage{bbm}
\usepackage{booktabs}
\usepackage[ruled,vlined,noend,linesnumbered]{algorithm2e}
\usepackage{algorithmic}
\usepackage{array}
\usepackage{enumitem}
\usepackage{makecell}
\newcolumntype{L}[1]{>{\raggedright\let\newline\\\arraybackslash\hspace{0pt}}m{#1}}
\newcolumntype{C}[1]{>{\centering\let\newline\\\arraybackslash\hspace{0pt}}m{#1}}
\newcolumntype{R}[1]{>{\raggedleft\let\newline\\\arraybackslash\hspace{0pt}}m{#1}}
\usepackage{multicol}
\usepackage{multirow}
\usepackage{xcolor}
\usepackage[hidelinks,colorlinks = true,
linkcolor = blue,
urlcolor  = blue,
citecolor = blue,
anchorcolor = blue,
bookmarks=false]{hyperref}
\usepackage{tabularx}

\newcommand{\cnf}{CNF}
\newcommand{\sygus}{SyGuS}

\newcommand{\zthree}{Z3}
\newcommand{\smt}{SMT}

\newcommand{\sat}{SAT}

\newcommand{\qbf}{QBF}
\newcommand{\dqbf}{DQBF}
\newcommand{\cadet}{CADET}
\newcommand{\manthan}{Manthan}

\newcommand{\depqbf}{DepQBF}
\newcommand{\dqbdd}{DQBDD}
\newcommand{\dcaqe}{DCAQE}

\newcommand{\cvcfour}{CVC4}

\newcommand{\dryadsynth}{DryadSynth}
\newcommand{\eusolver}{EUSolver}
\newcommand{\esolver}{ESolver}
\newcommand{\stochpp}{Stochpp}
\newcommand{\symbolic}{Symbolic}

\newcommand{\cvcfourcex}{CVC4$_{cex}$}
\newcommand{\cvcfourenum}{CVC4$_{enum}$}

\newcommand{\Signature}{\ensuremath{\mathsf{CallSigns}}}

\title{Program Synthesis as Dependency Quantified Formula Modulo Theory\thanks{The open source tool is available at \url{https://github.com/meelgroup/DeQuS}}}

\institute{}
\author{Priyanka Golia\inst{1,2} \and  Subhajit Roy\inst{1} \and Kuldeep S. Meel\inst{2}}
\authorrunning{Golia, Roy, and Meel}
\institute{ Computer Science and Engineering, Indian Institute of Technology Kanpur, India  \and School of Computing, National University of Singapore, Singapore\\
}

\begin{document}

\maketitle

\begin{abstract}
	Given a specification $\varphi(X,Y)$ over inputs $X$ and output $Y$, defined over a background theory $\mathbb{T}$, the problem of program synthesis is to design a program $f$ such that $Y=f(X)$ satisfies the specification $\varphi$. Over the past decade, syntax-guided synthesis (SyGuS) has emerged as a dominant approach for program synthesis where in addition to the specification $\varphi$, the end-user also specifies a grammar $L$ to aid the underlying synthesis engine. This paper investigates the feasibility of synthesis techniques without grammar, a sub-class defined as $\mathbb{T}$-constrained synthesis. 
	
	We show that $\mathbb{T}$-constrained synthesis can be reduced to DQF($\mathbb{T}$), i.e., to the problem of finding a witness of a Dependency Quantified Formula Modulo Theory. When the underlying theory is the theory of bitvectors, the corresponding DQF(BV) problem can be further reduced to  Dependency Quantified Boolean Formulas (DQBF). We rely on the progress in DQBF solving to design DQBF-based synthesizers that outperform the domain-specific program synthesis techniques, thereby positioning DQBF as a core representation language for program synthesis.  Our empirical analysis shows that $\mathbb{T}$-constrained synthesis can achieve significantly better performance than syntax-guided approaches. Furthermore, the general-purpose DQBF solvers perform on par with domain-specific synthesis techniques.
\end{abstract}

\section{Introduction} \label{sec:introduction}

The past three decades have been witness to the so-called {\em NP revolution} aided by the development of sophisticated heuristics for SAT solving leading to modern solvers being able to handle formulas involving millions of variables~\cite{MLM09}. The progress in SAT solving has led to effort for problems that lie in complexity classes {\em beyond NP} such as Quantified Boolean Formulas (QBF)~\cite{ACGKS18,LE17,R19,RTRS18}, Max-SAT~\cite{MML14}, model or MUS counting \cite{SRSM19,SGM20,BM20}, and the like. 

Of particular interest to us is the development of algorithmic approaches for QBF from theoretical and practical perspectives~\cite{ACGKS18,LB10,LE17,R19,RT15,RTRS18}. These improvements have paved the way for studying variants of QBF, which are harder from complexity-theoretic perspective. One such variant of interest is \textit{Dependency Quantified Boolean Formulas (DQBF)}, which generalizes the well known  notion of Quantified Boolean Formulas (QBF) by allowing explicit specification of dependency for existentially quantified variables, also known as {\em Henkin quantifiers}~\cite{H61}. The expressiveness of DBQF comes at the cost of the hardness from a complexity-theoretic perspective: in particular, DQBF is NEXPTIME-complete~\cite{PRA01}.  Nevertheless, as noted by Ganian et al.~\cite{GPSS20}, motivated by the progress in QBF solving, the past few years have seen a surge of interest from diverse viewpoints such as the development of DQBF proof systems, the study of restricted fragments to development of efficient DQBF solvers~\cite{LE17,RTRS18,TR19}. In this work, we will focus on DQBF and its generalization, Dependency Quantified Formulas modulo Theory, henceforth referred to as DQF($\mathbb{T})$.
 
A crucial ingredient in the {\em NP revolution } was the reduction of key problems such as planning~\cite{KS92} and bounded model checking~\cite{BCC+99} to SAT. Such reductions served as a rich source of practical instances, and at the same time, planning and bounded model checking tools built on top of SAT achieved fruits of the progress in SAT solving and thereby leading to even wider adoption, and contributing to a virtuous cycle~\cite{MLM09}.
 
Our investigation in this paper is in a similar spirit: we focus on a key problem in programming languages, program synthesis, and investigate its relationship to DQF($\mathbb{T}$) and DQBF. Given a specification $\varphi(X,Y)$ over the set of inputs $X$ and the set of  outputs $Y$, the problem of program synthesis is to synthesize a program $f$ such that $Y=f(X)$ would satisfy the specification $\varphi$. The earliest work on synthesis dates back to Church~\cite{K32}, and the computational intractability of the problem defied development of practical techniques. A significant breakthrough was achieved with the introduction of Syntax-Guided Synthesis (SyGuS)~\cite{ABJMRSSSTU13} formulation wherein in addition to $\varphi$, the input also contains a grammar of allowed implementations of $f$. The grammar helps to constrain the space of allowed implementation of $f$, and therefore, it also allows development of techniques that can efficiently enumerate over the grammar. While grammar has also been used as an implicit specification tool for few selected applications, it is mainly used to aid the underlying solver by constraining the search space.~\cite{ABJMRSSSTU13,ARU17,BCDHJK11}. Often, the end user is primarily concerned with any function that can be expressed using a particular theory $\mathbb{T}$. For the sake of clarity, we use the term {\em $\mathbb{T}$-constrained} synthesis\footnote{defined formally in Section~\ref{sec:prelims}} to characterize such class of synthesis problems. Two observations are in order: first, $\mathbb{T}$-constrained synthesis is a subclass of SyGuS, i.e., every instance of $\mathbb{T}$-constrained synthesis is also an instance of SyGuS. Of particular interest is the recent work in the development of specialized algorithms focused on $\mathbb{T}$-constrained synthesis, e.g.,  counterexample-guided quantifier instantiation algorithm in \cite{RDKTB15}. Secondly, recent studies have also highlighted that for a wide variety of applications, the usage of grammar is solely for the purpose of aiding solver efficiency, and as such have advocated usage of more expressive grammars for a given SyGuS instance~\cite{PMS19}.

The primary contribution of our work is establishing a connection between Theory-constrained synthesis and DQF($\mathbb{T})$. In particular, our work makes the following contributions:

\begin{description}
  	
  	 \item[From $\mathbb{T}$-constrained synthesis to DQF($\mathbb{T}$)] We present an reduction of $\mathbb{T}$-constrained synthesis to  DQF($\mathbb{T})$. DQF($\mathbb{T})$ lifts the notion of DQBF from the Boolean domain to general Theory $\mathbb{T}$. We view the simplicity of the reduction from $\mathbb{T}$-constrained synthesis to DQF$(\mathbb{T})$ as a core strength of the proposed approach.   %

  	\item[Efficient $\mathbb{T}$-constrained synthesizers for $\mathbb{T}$=bitvectors] The reduction to DQF($\mathbb{T})$ opens up new directions for further work. As a first step, we focus on the case when the $\mathbb{T}$ is restricted to bitvector theory, denoted by BV. We observe that the resulting DQF(BV) instances can be equivalently specified as a DQBF instance. We demonstrate that our reduction to DQBF allows us to simply plug-in the state of the art DQBF solvers~\cite{FKB14,GWRSSB15,RT15,J20}.
  	
  	\item[The Power of DQBF]  The remarkable progress in DQBF over the past few years is evident in our observation that DQBF-based synthesizers perform significantly better than domain-specific techniques that focus on utilization of the grammar to achieve efficiency. Our investigations were motivated by the classical work of Kautz and Selman~\cite{KS92} that showcased the power of general purpose SAT solvers in achieving advances over domain specific techniques. In this spirit, our work showcases the power of reducing synthesis to DQF($\mathbb{T}$). One question that we did not address so far is the choice of DQF$(\mathbb{T})$ over $\mathbb{T}$-constrained synthesis. To this end,  we again focused on the case when $\mathbb{T}$=BV, and we transform DQBF benchmarks to program synthesis, and  perform a comparison of program synthesis techniques vis-a-vis DQBF techniques. We observe that DQBF techniques significantly outperform the program synthesis techniques for these benchmarks; these results highlight the versatility of DQBF techniques and provide evidence in support of our choice of DQBF as the representation language.
  
  \item[Role of Grammars] Since DQBF-based synthesis techniques perform better than techniques that rely on grammar for efficiency, we would like to put forth the thesis of the usage of grammar as a specification tool rather than to guide the search strategy of the underlying synthesis engines, i.e., evolution of syntax-guided synthesis to syntax-constrained synthesis.

  \end{description}

The rest of the paper is organized as follows: we first introduce the notation and preliminaries in Section~\ref{sec:prelims}. We then discuss the related work in Section~\ref{sec:relatedwork}. In Section~\ref{sec:overview} we present a reduction of $\mathbb{T}$-constrained synthesis to DQF($\mathbb{T}$), then we talk about the specific case of $\mathbb{T}$=BV. We then describe the experimental methodology and discuss results in Section~\ref{sec:experiments}. Finally, we conclude in Section~\ref{sec:conclusion}.

\section{Notation and Preliminaries} \label{sec:prelims}
We will focus on the typed first order logic formulas associated with a background theory $\mathbb{T}$. We assume (without loss of generality) that function arguments are variables\footnote{we can achieve such a representation by simply replacing every term $t$ in the argument of a function $f$ with a variable $x$ and conjuncting $\varphi$ with a constraint $x = t$}. We refer to the function and its (ordered) list of arguments at an invocation (within $\varphi$) as its \textit{call signature}. The set of all call signatures of a function symbol $f$ in $\varphi$ is referred by ${\Signature}(f)$. Note that the number of invocations of a function may not match $|\Signature(f)|$. For example, the following formula $\varphi$, has $4$ invocations of $f$ while \Signature(f) = 
$\{ \langle a,b \rangle, \langle b,c \rangle, \langle b,a \rangle \}$. Note that $\langle a,b \rangle$ and $\langle b,a \rangle$ are considered as two different $\Signature$ of $f$.

$$\varphi: \forall a,b,c \ \exists f \ f(a,b) \land f(b,c) \land f(b,a) \land f(a,b)$$

If every invocation of $f$ in $\varphi$ has an identical argument-list, then $|\Signature(f)| = 1$, and we refer to $\varphi$ as a \textit{single-callsign} instance. Otherwise, $\varphi$ is a \textit{multiple-callsign} instance. We use $args(f)$ to get the argument lists of function $f$, and ${f(args)}$ to represents the invocation of $f$ with its \emph{args}.
While from the point of view of expressiveness, multiple functions can be reduced to the case of a single function, the  performance of synthesis tools is affected by the number of functions to be synthesized. 

We use BV for the bitvector theory. We use the lower case letters with subscripts for a variable, and the upper case letters for the set of variables. A \textit{literal} is either a variable or its negation, and a \textit{clause} is a disjunction of literals. A \textit{conjunctive normal form} ({\cnf}) formula is a conjunction of clauses.

We say that  $\varPhi$ is a Dependency-Quantified Boolean Formula (\dqbf) if $\varPhi$ can be represented as: 
$\forall X\,.\exists^{H_1} y_1.\,\ldots, \exists^{H_m} y_m.\,\varphi(X,Y)$, where $\varphi$ is a quantifier-free propositional formula and $H_i \subseteq X$ such that the variables $y_i$ can depend only on the variables of $H_i$. We refer to $\exists^{H_i}$ as a \textit{Henkin quantifier}. We also use the notation $\exists^{H}V$ to represent that every variable of the set $V$ can depend only on the variables of $H$. If $\forall_{i \in [1,m]} H_i = X$, then $\varPhi$ is considered as a 2-Quantified Boolean Formula (2-QBF).

\paragraph{Henkin Functional Synthesis:} For a given Boolean relational specification between a set of inputs $X$ and a set of outputs Y as $\exists^{H_1} y_1,\ldots, \exists^{H_m} y_m.\,\varphi(X,Y)$ where $X = \{ x_1,\ldots, x_n\}$, $Y = \{y_1,\ldots,y_m\}$ and each $H_i \subseteq X$, synthesize a function vector $\boldsymbol{f} = \langle f_1(H_1), \ldots, f_m(H_m) \rangle$, such that $y_i \leftrightarrow f_i(H_i)$ and  $\exists^{H_1} y_1\,\ldots \exists^{H_m} y_m.\,\varphi(X,Y) \equiv \varphi(X,\boldsymbol{f})$, where $\varphi(X,\boldsymbol{f})$ represent the formula $\varphi$ upon substituting $y_i$ with corresponding function $f_i$.
The function vector $\boldsymbol{f}$ is also called \textit{Henkin function vector} and each $f_i$ is called \textit{Henkin function}.

A DQBF solver search for an existences of such Henkin function vector $\boldsymbol{f}$ such that $\varphi(X,\boldsymbol{f})$ becomes tautology. If such a function vector $\boldsymbol{f}$ exists, DQBF solver outputs True, else False. A certifying DQBF solver also outputs  $\boldsymbol{f}$ in addition to True/False.

\section{Related Work} \label{sec:relatedwork}

From a theoretical perspective, the satisfiability problem of Dependency Quantified Boolean formulas ({\dqbf}) is known to be NEXPTIME-complete~\cite{PRA01}. Despite the theoretical hardness, the past decades have been witness to substantial development in both {\qbf} and {\dqbf} solving~\cite{R19,RT15,J20,TR19}. Frohlich et al.~\cite{FKB12} proposed the first DPLL based algorithm to lift the {\qbf} solving to {\dqbf}; Tentrup and Rabe~\cite{TR19} followed the same direction, and proposed the idea of using clausal abstraction for {\dqbf} solving. In terms of {\qbf} solvers, the series of work by Rabe et al.~\cite{R19,RT15} showed that the idea of using incremental determination not only improves the state-of-the-art for {\qbf} solving, but also can be used for Skolem functional synthesis. In a separate line of work Jiang~\cite{J09} introduced \emph{Craig interpolation-based} Skolem function synthesis approach. Subsequently, CEGAR-based approaches have also been proposed~\cite{AACKRS19,ACGKS18}. Fried, Tabajara and Vardi~\cite{FTV16} combined RBDD-based approach with CEGAR for Skolem function synthesis. Recently, Golia, Roy and Meel~\cite{GRM20} proposed a data-driven approach for Skolem functional synthesis.

The origins of program synthesis trace back to 1930s. The work on constructive mathematics~\cite{K32} initiated the idea of constructing interpretable solutions with proofs by composing solutions of smaller sub-problems, followed up by deductive synthesis strategies~\cite{G81,MW71}. Alur et al.~\cite{ABJMRSSSTU13} proposed the idea of using grammar for syntax-guided synthesis; they showed that using grammar would result in efficient program synthesis with more interpretable programs~\cite{ABJMRSSSTU13,ARU17,URDMA13}. Another line of work is to build the synthesizer on top of  {\sat}/{\smt} solver~\cite{RDKTB15}. {\cvcfour}~\cite{RDKTB15} is the first synthesis engine implemented inside an {\smt} solver, it extracts the desired functions from the unsatisfiability proofs of the negated form of synthesis conjectures. 

In recent years, there has been work on understanding the impact of provided grammar on the performance of existing tools. Padhi et al.~\cite{PMS19} empirically showed that with the increase of expressiveness of the provided grammar, the performance of the tools deteriorates significantly. Recently, Kim et al.~\cite{KHR21} introduced \emph{semantic guided synthesis}, where the user is allowed to specify both syntax and semantics.

Prior works have identified the use of {\dqbf} solving for reactive and bounded synthesis~\cite{FFRT17,GWRSSB15,TR19}. Tentrup and Rabe~\cite{TR19} demonstrates that the usage of clausal abstraction for {\dqbf} leads to performance improvements for the benchmarks arising from reactive synthesis. In this context, our reduction from program synthesis for BV programs to {\dqbf} further reinforces the centrality of {\dqbf} as a representation language.

\section{Synthesis via DQF($\mathbb{T}$)} \label{sec:overview}

Let us first discuss the well-known formulation of syntax-guided synthesis (\sygus). In this formulation, the constraints over the functions to be synthesized are specified in the vocabulary of a given background theory $\mathbb{T}$ along with the function symbols. Notice that the background theory specifies the domain of values for each variable type along with the interpretation for the function(s) and predicate symbols in the vocabulary. 

\begin{definition}[\cite{ABJMRSSSTU13}]
	Given a background theory $\mathbb{T}$, a set of typed function symbols $\{f_1, \ldots f_k\}$, a specification $\varphi$ over the vocabulary of $\mathbb{T} \cup \{f_1, f_2, \ldots f_k\}$, a set of expressions $\{L_1,\ldots L_k\}$ over the vocabulary of~$\mathbb{T}$ such that $L_i$ is of the same type as $f_i$, the problem of syntax-guided synthesis (SyGuS) is to find a set of  expressions $\{e_1 \in L_1 , e_2 \in L_2, \ldots e_k \in L_k \}$ such that 
	the formula $\varphi[f_1/e_1, f_2/e_2, \ldots f_k/e_k]$ is valid modulo $\mathbb{T}$.
	Note that $\varphi[f_i/e_i]$ denotes the result of substitution of $f_i$ with expression $e_i$ such that the bindings of inputs to $f_i$ is ensured. 
\end{definition} 

In this work, we are interested in the subclass of SyGuS where $L_i$ corresponds to the \textit{complete} vocabulary of $\mathbb{T}$; we call such a class as $\mathbb{T}$-constrained synthesis, defined formally below:

\begin{definition}
	Given a background theory $\mathbb{T}$, a set of typed function symbols $\{f_1, \ldots f_k\}$, a specification $\varphi$ over the vocabulary of $\mathbb{T} \cup \{f_1,\ldots f_k\}$, the problem of $\mathbb{T}$-constrained synthesis is to find the set of  expressions $\{e_1, e_2 , \ldots e_k \}$ defined over vocabulary of $\mathbb{T}$ such that 
	the formula $\varphi[f_1/e_1, f_2/e_2, \ldots f_k/e_k]$ is valid modulo $\mathbb{T}$. %
	
\end{definition}
We propose the reduction of $\mathbb{T}$-constrained synthesis to DQF($\mathbb{T}$), i.e, to the problem of finding a witness of a dependency quantified formula modulo theory.

\subsection{$\mathbb{T}$-Constrained Synthesis to DQF($\mathbb{T}$)} 

As remarked in Section~\ref{sec:introduction}, a key strength of the reduction is its simplicity. Algorithm~\ref{algo:single-reduction} formalizes the desired reduction of $\varphi$ to DQF($\mathbb{T}$) formulation where $\varphi$ is a specification over the vocabulary of background theory $\mathbb{T}$ with a set of typed function symbols $\{f_1, f_2, \ldots f_m\}$ such that for all $f_i$, $|{\Signature}(f_i)| = 1$. The important point to note is that the Henkin quantifiers must be carefully constructed so that each $f_i$  depends only on the set of variables that appear in its argument-list.

\begin{algorithm}[h]
	\DontPrintSemicolon
	\SetAlgoLined
	\KwIn{ A background theory $\mathbb{T}$, a set of typed function symbols $\{f_1, f_2, \ldots f_m\}$, a specification $\varphi$ over the vocabulary of $\mathbb{T}$}
	
	Let $X = \bigcup_{f_i} \{h \mid  h \in \Signature(f_i) \}$\label{algo:single-reduction:line:declare}\;
	
	Substitute every invocation of $f_i$ with a fresh variable $y_i$ in $\varphi$\label{algo:single-reduction:line:freshvar}\;
	
	Define $H_{i} = Set(h)$ as $\{h|h \in \Signature(f_i)\}$\label{algo:single-reduction:line:define-hi}\;
	
	\KwOut{$\forall X \exists^{H_1} y_1.\ \exists^{H_2} y_2\ \dots \exists^{H_{m}} y_{m} \varphi (X,Y)$}
	
	\caption{Reducing single-callsign instance $\varphi$ to DQF($\mathbb{T}$)\label{algo:single-reduction}}
\end{algorithm}

Now, let us turn our attention to the case when there exist a function $f_i$ such that $|{\Signature}(f_i)| > 1$. In such cases, we pursue a Ackermannization-style technique
that transforms $\varphi$ into another specification $\hat{\varphi}$ such that every function $f_i$ in $\hat{\varphi}$ has $|{\Signature}(f_i)| = 1$ (Algorithm~\ref{algo:reduction-hatvarphi}). Note that this transformation allows the subsequent use of Algorithm~\ref{algo:single-reduction} with $\hat{\varphi}$ to complete the reduction to DQF($\mathbb{T}$). The proposed transformations in Algorithm~\ref{algo:reduction-hatvarphi} are linear in the size of the formula like the transformation introduced in~\cite{R17}, however Algorithm~\ref{algo:reduction-hatvarphi} introduces lesser number of new variables.

The essence of Algorithm~\ref{algo:reduction-hatvarphi} is captured in the following two transformations:
\begin{description}
	\item \textbf{(Line~\ref{algo:reduction-hatvarphi:line:sub})} We substitute instances of every call signature of $f_i$ with fresh function symbols $f^j_i$ (that corresponds to the $j^{th}$ call signature of $f_i$). This reduces the formula from multiple-callsign to a single-callsign instance.
	\item \textbf{(Line~\ref{algo:reduction-hatvarphi:line:constraint})} Introduction of an additional constraint for each $f_i$ that forces all the functions $f_i^j$ (introduced above) to mutually agree on every possible instantiation of arguments. Specifically, it introduces a fresh function symbol $f_i^{l_i}$ and a set of fresh variables $z^{i}_1,\dots,z^{i}_n\in Z_i$ such that, for all argument lists $args(f_i^{j})$:
	$$(args(f_i^{j}) = Z_i ) \implies f_i^{j}(args) = f_i^{\ell_i}(Z_{i})$$ 
        where $j \in [0\dots l_{i-1}]$.
\end{description}

\begin{algorithm*}
	\DontPrintSemicolon
	\SetAlgoLined
	\KwIn{A background theory $\mathbb{T}$, a set of typed function symbols $\{f_1, f_2, \ldots f_m\}$, a specification $\varphi$ over the vocabulary of $\mathbb{T}$ such that $\ell_i = |\Signature(f_i) |$}

	\For{$i=1$ to $m$}
	{
		\If{$|\Signature(f_i)| > 1$}
		{
			Add a fresh (ordered) set of variables $Z_{i}$ such that $|Z_{i}| = |\Signature(f_i)[0]|$ \;
			
			\For{$j \in [0\dots (\ell_i - 1)]$}
			{
				
				Replace every $f_i$ whose $args(f_i) = \Signature(f_i)[j]$  with $f_i^{j}$\label{algo:reduction-hatvarphi:line:sub}\;
				
				Add constraint $(args(f_i^{j}) = Z_i ) \rightarrow f_i^{j}(args) = f_i^{\ell_i}(Z_{i})$ to $\varphi$ \label{algo:reduction-hatvarphi:line:constraint}\;
				
			}
			$\Signature(f_i) \gets \Signature(f_i) \cup \{Z_{i}\}$\;
		}
		
	}
	\KwOut{A set of typed function symbols $\{ f_1^{0}, f_1^{2}, \ldots f_1^{\ell_1}, \ldots, f_m^{0} \ldots f_m^{\ell_m} \}$, a specification $\hat{\varphi}$ 
		over the vocabulary of $\mathbb{T}$ such that $\forall i,j$ we have  $|\Signature(f_i^{j})| = 1$}
	\caption{Reducing multiple-callsign to single-callsign instance\label{algo:reduction-hatvarphi}}
\end{algorithm*}

\subsection{When $\mathbb{T}$ is bitvector (BV):}
\label{sec:t=bitvector}
When the specification $\varphi(X,Y)$ is in BV, we solve the problem in the following steps:

\begin{enumerate}
    \item \textit{Reduction to single-instance:} If $\varphi(X, Y)$ is a multiple-callsign instance, we use Algorithm~\ref{algo:reduction-hatvarphi} to covert it to a single-callsign instance $\hat{\varphi}(\hat{X},\hat{Y})$.
    \item \textit{Reduction to DQF(BV):} We use Algorithm~\ref{algo:single-reduction} to generate the DQF(BV) instance of $\hat{\varphi}(\hat{X},\hat{Y})$ as $\forall \hat{X} \exists^{H_1} \hat{y}_1.\ \exists^{H_2} \hat{y}_2\ \dots \exists^{H_{m}} \hat{y}_{m} \hat{\varphi} (\hat{X},\hat{Y})$.
    \item \textit{Solving DQF(BV):} We solve the DQF(BV) instance by compiling it down to a DQBF instance, thereby allowing the use of off-the-shelf DQBF solvers. We detail this step in the following discussion.
\end{enumerate}

As the first step to DQBF compilation, we perform {\em bit-blasting} over ${\hat{\varphi}}$ to obtain $\hat{\varphi}'$. 
\begin{multline}\label{eq:qbf}
 \forall \hat{X} \exists^{H_1} \hat{y}_1\ \ldots \exists^{H_{m}} \hat{y}_{m} \hat{\varphi} (\hat{X},\hat{Y}) \equiv  \forall {{X}'} \exists^{{{X}'}} V.\ \exists^{{H}_1^{'}} {{Y_1^{'}}} \ldots \exists^{{H}_m^{'}} {Y_m^{'}}  {{\varphi'}}({{X}'},{{Y}'})
\end{multline}

where, ${X'}, {Y_i^{'}}, H_{i}^{'}$ are the (bit-blasted) \textit{sets} of propositional variables mapping to the bitvector variables $\hat{X}, \hat{y_i}, {H_{i}}$ respectively. Furthermore, $V$ is the set of auxiliary variables introduced during bit-blasting. The auxiliary variables can be allowed to depend on all the input variables ${X'}$. From an efficiency perspective, one can record the corresponding Henkin functions for the auxilary variables during bit-blasting process; we leave such optimizations to future work. Our current framework simply employs off-the-shelf SMT solvers for the bit-blasting.

As the formula on the right-hand side in Eq.~\ref{eq:qbf} is an instance of DQBF, we can simply invoke an off-the-shelf {\em certifying DQBF} solvers to generate the Henkin functions for ${Y'}$. A careful reader will observe that the Henkin functions corresponding to ${y}'_i$ variables will be constructed in the propositional theory and not in BV, but note that one can simply convert a formula in propositional theory into an equivalent formula in bitvector theory defined over $\hat{X}, \hat{y}_1, \ldots \hat{y}_m$ with only a linear size increase in the representation. 

While theoretical analysis is not the primary focus of this work, we observe that our reduction to DQBF allows one to lift the recently obtained results for DQBF to obtain Fixed Parameter Tractability (FPT) analysis for synthesis~\cite{GPSS20}. We leave a detailed theoretical analysis to future work.

\section{Experimental Evaluation} \label{sec:experiments}

The objective of our experimental evaluation was to study the feasibility of solving BV-constrained synthesis via the state-of-the-art {\dqbf} solvers. To this end, we perform an evaluation over an extensive suite of benchmarks and tools, which we describe in detail below. 

\subsection{Experimental Setup}

\subsubsection{Tools under evaluation}
Given a program synthesis instance defined over BV, we sought to compare three different possibilities: executing a {\sygus} tool, executing a BV-constrained synthesis, and a DQBF-augmented synthesis framework. To this end, we experimented with the following tools (Table~\ref{tab:summary-tools}): 

\begin{description}
	\item[Syntax-Guided Synthesis:] {\sygus} tools for bitvector theory spanning symbolic, stochastic and enumerative solvers, like {\cvcfour}~\cite{BCDHJK11}, {\eusolver}~\cite{ARU17}, {\esolver}~\cite{URDMA13}, {\stochpp}~\cite{ABJMRSSSTU13}, {\symbolic} Solver~\cite{ABJMRSSSTU13}, {\dryadsynth}\cite{HQSW20}. Since we employ multiple versions of {\cvcfour}, we will refer to the {\sygus}-based variant of {\cvcfour} as {\cvcfourenum}.
	
	\item[BV-Constrained Synthesis:] Observe that every SyGuS tool can be transformed into a theory-constrained tool by a simple pipeline that rewrites the grammar of the input instance to be the entire vocabulary of the corresponding theory. Our empirical evaluation, however, revealed that such a transformation is ineffective, and the tools under transformation consistently perform worse than their corresponding versions under syntax-guided mode. In addition to the SyGuS-based BV-constrained tools, we employ the state of the art synthesis engine {\cvcfour}~\cite{RDKTB15} under the counter-example-guided quantifier instantiation mode, which can be viewed as a native approach to BV-constrained synthesis. We refer {\cvcfourcex} to denote {\cvcfour} invoked with counter-example-guided quantifier instantiation.
	
	\item[{\dqbf}-based Synthesis:] The set of underlying DQBF solvers that we have employed in DQBF-based synthesis framework span {\cadet}~\cite{R19}, {\manthan}~\cite{GRM20}, {\depqbf}~\cite{LB10}, {\dcaqe}~\cite{TR19} and {\dqbdd}~\cite{J20}. A careful reader might observe that {\cadet} is a QBF solver, and {\manthan} is a Skolem function synthesizer, i.e., they can only handle the special case when the existentially quantified variables are allowed to depend on all the universally quantified variables.  
\end{description} 

\begin{table}[h]
	\centering
    \caption{\label{tab:summary-tools}Tools used in our evaluation}
	\begin{tabular}{C{3cm} C{1.5cm} C{2cm}}
		\midrule
		 Syntax-Guided  & BV-Constrained & {\dqbf}-based \\ \hline
		
		\makecell{{\cvcfourenum},{\esolver} \\ {\eusolver},\\ {\dryadsynth}, {\stochpp} \\  {\symbolic} Solver } 
		
		&  \makecell{\cvcfourcex}
			 
		& \makecell{{\cadet}, {\manthan}\\ {\depqbf}, {\dqbdd}\\ {\dcaqe}}
		\\
		\hline
	\end{tabular}
\end{table}

\subsubsection{Benchmarks}
The benchmark suite consisted of instances from two sources: SyGuS competition and QBF competition. 
 We use $645$ general-track bitvector~(BV) theory benchmarks from SyGuS competition 2018, 2019~\footnote{\url{https://sygus.org/}} to evaluate the performance of {\sygus} tools.  
To employ our DQBF-based framework, we used {\zthree}~\cite{DB08} to convert the instances from SyGuS to DQBF. 
Furthermore, we considered $609$ {\qbf} benchmarks from QBFEval competition 17,18\footnote{\url{http://www.qbflib.org/index_eval.php}}, disjunctive decomposition and arithmetic set~\cite{AACKRS19,R19} and converted them to {\sygus} instances. We considered each propositional variable as a bitvector of size $1$, and allow the synthesized function to depend on all the universally quantified variables. 
The associated grammar for these benchmarks is the entire BV-vocabulary.

\subsubsection{Implementation and Setup}
All our experiments were conducted on a high-performance computer cluster with each node consisting of a E5-2690 v3 CPU with $24$ cores and $96$GB of RAM, with a memory limit set to $4$GB per core. All tools were run in a single-threaded mode on a single core with a timeout of $900$s.

\subsection{Objectives and Summary}

The primary objective of empirical evaluation is to show how general purpose DQBF techniques can be transformed to effective synthesis engines.  In particular, our empirical evaluation sought answers to the following questions:

\begin{description}
	\item[RQ1: Utility of Syntax for Efficiency] Can theory constrained synthesis work as well as Syntax-Guided Synthesis? 
	\item[RQ2: Domain Specific vs DQBF] Can general purpose DQBF solver-based synthesis framework   match the efficiency of domain specific synthesis tools?
	\item [RQ3: Efficacy of DQBF as Representation Language] Are BV-constrained synthesis an efficient approach to DQBF solving? 
\end{description}

\subsubsection{Summary of Results}

 Table~\ref{tab:summary-details} represents the instances solved by the virtual best solver for SyGuS, BV constrained, and DQBF tools. The first row of Table~\ref{tab:summary-details} lists the number of instances solved by different synthesis techniques: {\sygus} based, BV-constrained synthesis, and {\dqbf} based synthesis for {\sygus} instances, and the second row represents the same for {\dqbf} instances.
 
   \begin{table}[h]
   	\centering
   	\captionof{table}{\label{tab:summary-details}Number of {\sygus} and {\dqbf} instances solved using different techniques. Timeout 900s.}
   	\begin{tabularx}{\textwidth}{XXXXX}
   	\\	\toprule
   		& {Total}	& {SyGuS-tools}  & {BV-constrained} & {{\dqbf}-based}  \\ \midrule
   		
   		{\sygus}  & 645 & 513 & 606 & 610  \\ 
   		{\dqbf}  & 609 & - & 2 & 276 \\
   		\midrule
   		All & 1254 & 513 & 608 & 886 \\
   		\bottomrule
   	\end{tabularx}
   	
   \end{table}  
  
 As shown in Table~\ref{tab:summary-details}, with syntax guided synthesis, we could synthesize the functions for $513$ out of $645$ {\sygus} instances only, whereas, with BV-constrained synthesis, we could solve $606$ such instances. Surprisingly,  BV-constrained synthesis performs better than the syntax-guided synthesis. 
  
 Table~\ref{tab:summary-details} also shows that the {\dqbf} based synthesis tools perform similar to BV-constrained synthesis tools for {\sygus} instances; this provides strong evidence that the general purpose {\dqbf} solvers can match the efficiency of the domain specific synthesis tools. {\cvcfourcex} was the only tool we could find in the
 BV constrained category. However, Table~\ref{tab:summary-details} and~\ref{tab:sygus-benchmarks}
 show that even {\cadet} and {\manthan} alone solve
 879 and 868 instances respectively. Therefore, the
 advantage is not solely due to portfolio.  
 
 Also, Table~\ref{tab:summary-details} shows that BV-constrained synthesis tools perform poorly with {\dqbf} as representation language providing support for the efficacy of {\dqbf} as a representation language.

\subsection{Detailed Analysis}

\subsubsection{Performance Analysis for {\sygus} Instances}
\label{sec:experiments:sygus}
We used {\sygus} instances to evaluate the performance with different synthesis strategies: {\dqbf} based synthesis, BV-constrained synthesis and syntax-guided synthesis. We further divide the {\sygus} instances into four sub-categories: single-function-single-callsign, single-function-multiple-callsign, multiple-function-single-callsign, multiple-function-multiple-callsign.

\begin{table}
	\centering
	\caption{SyGuS tools performance with Single-Function Single-CallSign instances}~\label{tab:sygus-sfsc}
	\begin{tabularx}{\textwidth}{XXXXXXX} \\ \toprule
	&{\dryadsynth} & {\stochpp} & {\symbolic} &{\esolver} & {\eusolver} & {\cvcfourenum} \\ \midrule  
Total:609 &	15 & 39 & 108 & 151 & 236 & 488 \\ \bottomrule
	\end{tabularx}
\end{table}

Since the single-function-single-callsign instances can be converted into {\qbf} instances (instead of {\dqbf}), we employ the state-of-the-art {\qbf} solvers {\cadet}~\cite{R19} and {\depqbf}~\cite{LE17}, {\manthan}~\cite{GRM20} for these instances. With respect to RQ1, it turns-out that {\cvcfour} performs better with BV-constrained synthesis than with syntax-guided synthesis, as for single-function-single-callsign instances, {\cvcfourcex} could solve $602$ instances whereas {\cvcfourenum} could solve only $488$. However, as shown in Table~\ref{tab:sygus-sfsc}, {\cvcfourenum} outperforms the other state-of-the-art {\sygus} tools significantly. The second best {\sygus} tool was {\eusolver} which could solve only $236$ instances.

\begin{table}
	\centering
	\caption{DQBF solvers performance with Single-Function Single-CallSign instances}~\label{tab:dqbf-sfsc}
	\begin{tabularx}{0.80\textwidth}{XXXX} \\ \toprule
		& {\depqbf} & {\manthan} & {\cadet}  \\ \midrule  
		Total:609 &	564 & 592 & 605 \\ \bottomrule
	\end{tabularx}
\end{table}

Concerning RQ2, as shown in Table~\ref{tab:sygus-sfsc}, and~\ref{tab:dqbf-sfsc}, {\dqbf} solvers perform on par to domain specific synthesis tools, in-fact they perform slightly better for single-invocation-single-callsign category as {\cadet} could synthesize a function for $3$ more instances than {\cvcfourcex}. 

\begin{table}
	\centering
	\caption{\label{tab:sygus-benchmarks}The top three tools for each category are listed in the order of their performance, and the number (in bracket) represents the number of instances solved.}
	\begin{tabularx}{0.8\textwidth}{XXX} 
	\\	\toprule
		
		& Single-CallSign & Multiple-CallSign \\\midrule
		
		Single-Function &
		\makecell{\textcolor{blue}{Total instances: 609} \\ \textbf{{\cadet}(605)},\\ {\cvcfourcex}(602), \\{\manthan}(592)} 
		
		& \makecell{\textcolor{blue}{Total instances: 9}\\ \textbf{{\cvcfourenum}(8)}, \\ {\dqbdd}(2),\\{\dcaqe}(2) } \\ 
		
		\midrule
		
		Multiple-Functions & \makecell{\textcolor{blue}{Total instances: 7}\\ \textbf{{\cvcfourenum}(5)},\\{{\cvcfourcex}(4)}, \\ {\dqbdd}(3) }  &  \makecell{\textcolor{blue}{Total instances: 19}\\ \textbf{{\cvcfourenum}(12),}\\{\dqbdd}(0),\\ {\dcaqe}(0)} \\ 
		
		\bottomrule
		
	\end{tabularx}
\end{table}

Table~\ref{tab:sygus-benchmarks} represents the overall analysis for all four categories of {\sygus} instances. As shown in Table~\ref{tab:sygus-benchmarks}, {\dqbf} solvers and {\cvcfourcex}  have similar performance in terms of number of instances solved except for the multiple-functions-multiple-callsign category, and single-functions-multiple-callsign category.

\subsubsection{Performance Analysis for {\qbf} Instances}

We performed an experiment with the {\sygus} language as a representation language instead of {\qbf}. We considered {\qbf} instances instead of {\dqbf} as majority of our synthesis benchmarks could reduced to {\qbf}, and {\dqbf} is a generalization of {\qbf}. As {\cvcfourcex} performed the best amongst the different tools for BV-constrained synthesis for {\sygus} instances, we considered the {\cvcfourcex} to evaluate the performance over $609$ {\sygus} language representation of {\qbf} instances.

\begin{table}
	\centering
	\captionof{table}{\label{tab:qbf-benchmarks} Instances solved for BV-constrainted synthesis of {\qbf} benchmarks.
	}
	\begin{tabularx}{0.8\textwidth}{XXXXXX} \\ \toprule
		TO & {Total} & {\cvcfourcex} & {\depqbf} & {\cadet} & {\manthan} \\ \midrule
		
		900s	& 609 & 2 & 33 & 274 & 276  \\
		7200s & 609 & 2 & 39  & 280  & 356 \\
		\bottomrule
	\end{tabularx}
\end{table}

Table~\ref{tab:qbf-benchmarks} represents the performance analysis. We performed experiments with two settings: the same timeout ($900s$) as used for the tools in Section ~\ref{sec:experiments:sygus}, and a more relaxed timeout of $7200s$. With the $900s$ timeout, {\cvcfourcex} could solve only $2$ instances out of $609$ total instances, whereas, {\manthan} preformed the best by synthesizing the functions for $276$ instances. With the $7200s$ timeout, {\cvcfourcex} could not solve any new instances while {\manthan} solved another $80$ instances. Hence, BV-constrained synthesis is not an efficient approach for {\dqbf} solving, which answers RQ3.

\section{Conclusion}\label{sec:conclusion}
Syntax-guided synthesis has emerged as a dominant paradigm for program synthesis. Motivated by the impressive progress in automated reasoning, we investigate the usage of syntax as a tool to aid the underlying synthesis engine. To this end, we formalize the notion of $\mathbb{T}$-constrained synthesis, which can be reduced to DQF$(\mathbb{T})$. We then focus on the special case when $\mathbb{T}=BV$. The corresponding BV-constrained synthesis can be reduced to DQBF, highlighting the importance of the scalability of DQBF solvers. It is important to acknowledge that not every application of SyGuS employs grammar as a tool to aid solver efficiency; grammar can also be used as a specification tool such as to ensure that the synthesized program does not leak information~\cite{ATGK19}. Since properties such as information flow leakage can also be expressed as hyper-properties, we hope that our results will motivate the study of synthesis formulation with richer specifications. 

Our empirical analysis shows that $\mathbb{T}$-constrained synthesis can achieve significantly better performance than syntax-guided approaches. Furthermore, the general purpose DQBF solvers perform on par with domain-specific synthesis techniques and thereby supporting the argument of viewing DQBF as a general purpose representation language for representation task. We believe that our results will motivate further research into DQBF; the rewards of which can be reaped by program synthesis tools. 

It is certainly worth remarking that our experimental results were presented for the case when $\mathbb{T}$ was restricted to BV, and therefore, a valid criticism to the above proposed hypothesis would be lack of evidence for theories beyond bitvectors. In this regard, an interesting direction of future research would be design of techniques for DQF($\mathbb{T}$).

\paragraph{Acknowledgment}
We are thankful to the anonymous reviewers for their constructive comments and suggestions that improved the final draft of the paper. We are beholden to Saswat Padhi, Andrew Reynolds, and Abhishek Udupa for responding to our queries regarding prior work.

This work was supported in part by National Research Foundation Singapore under its NRF Fellowship Programme [NRF-NRFFAI1-2019-0004 ] and AI Singapore Programme [AISG-RP-2018-005], and NUS ODPRT Grant [R-252-000-685-13]. The computational work for this article was performed on resources of the National Supercomputing Centre, Singapore: \url{https://www.nscc.sg}~\cite{nscc}.

\clearpage
\bibliographystyle{splncs04}
\bibliography{ref}
\end{document}